\documentclass[10pt,twocolumn,letterpaper]{article}

\usepackage{wacv}
\usepackage{times}
\usepackage{epsfig}
\usepackage{graphicx}
\usepackage{amsmath}
\usepackage{amssymb}

\usepackage{times}
\usepackage{epsfig}
\usepackage{psfrag}
\usepackage{graphicx}
\usepackage{amsmath}
\usepackage{amssymb}
\usepackage{mathrsfs}
\usepackage{subfigure}
\usepackage{enumerate}
\usepackage[lined,boxed,commentsnumbered]{algorithm2e}
\usepackage{multirow}
\usepackage{caption}
\usepackage{authblk}

\usepackage[pagebackref=true,breaklinks=true,letterpaper=true,colorlinks,bookmarks=false]{hyperref}

\wacvfinalcopy % *** Uncomment this line for the final submission

\def\wacvPaperID{****} % *** Enter the wacv Paper ID here

\ifwacvfinal\fi
\begin{document}

%%%%%%%%% TITLE
\title{Unsupervised Network Pretraining via Encoding Human Design }
\author[1]{Ming-Yu Liu\thanks{mliu@merl.com}}
\author[2]{Arun Mallya}%\thanks{arun.mallya@gmail.com}}
\author[1]{Oncel Tuzel}%\thanks{oncel@merl.com}}
\author[3]{Xi Chen}%\thanks{chenxistephen@gmail.com}}
\affil[1]{Mitsubishi Electric Research Labs (MERL), Cambridge MA, USA}
\affil[2]{University of Illinois, Urbana-Champaign, IL, USA}
\affil[3]{University of Maryland, College Park, MD, USA}
% \author{Ming-Yu Liu, Arun Mallya, Oncel Tuzel, Xi Chen\\
% Mitsubishi Electric Research Labs\\
% Cambridge, MA, USA\\
% {\tt\small mliu@merl.com}
% \and
% Arun Mallya\\
% University of Illinois\\
% Urbana-Champaign, IL, USA\\
% {\tt\small arun.mallya@gmail.com}
% \and
% Oncel Tuzel\\
% Mitsubishi Electric Research Labs\\
% Cambridge, MA, USA\\
% {\tt\small oncel@merl.com}
% \and
% Xi Chen\\
% University of Maryland\\
% College Park, MD, USA\\
% {\tt\small chenxistephen@gmail.com}
% }

\maketitle
\thispagestyle{empty}

\begin{abstract}
Over the years, computer vision researchers have spent an immense amount of effort on designing image features for the visual object recognition task. We propose to incorporate this valuable experience to guide the task of training deep neural networks. Our idea is to pretrain the network through the task of replicating the process of hand-designed feature extraction. By learning to replicate the process, the neural network integrates previous research knowledge and learns to model visual objects in a way similar to the hand-designed features. In the succeeding finetuning step, it further learns object-specific representations from labeled data and this boosts its classification power. We pretrain two convolutional neural networks where one replicates the process of histogram of oriented gradients feature extraction, and the other replicates the process of region covariance feature extraction. After finetuning, we achieve substantially better performance than the baseline methods.
\end{abstract}

\section{Introduction}

Deep learning methods are revolutionizing the field of visual object recognition. Starting from the breakthrough in image classification~\cite{Krizhevsky_KSH_NIPS12}, similar successes were achieved for several other computer vision tasks~\cite{Girshick_GDDM_CVPR14,Taigman_TYRW_CVPR14}. These have demonstrated that powerful feature representations can be learned from data automatically, out-dating traditional approaches based on hand-designed features. Following these successes, the focus of visual object recognition research has shifted from feature engineering to deep network design and optimization. One of the techniques developed for deep network optimization is pretraining, which helps when the amount of labeled data is limited.

Pretraining refers to the technique of initializing the network parameters with the ones learned from applying the network to solve a different task. Most of the existing pretraining approaches are based on unsupervised learning~\cite{hinton2006fast,Bengio_BLPL_NIPS,ranzato2006efficient}. These works pretrain the deep network in a layer-wise manner where a layer is pretrained after the preceding layer is pretrained. Each layer learns a non-linear transformation from the input to an intermediate representation which can be used to reproduce the input data. The reproducing capability suggests that the network encodes the main variation of the input data. Once the pretraining process is completed, the resulting network parameters are used to initialize the network, which is further optimized using labeled data; a process known as finetuning. It has been shown that the unsupervised pretraining guides the learning towards basins of attraction of minima that support better generalization capability~\cite{Erhan_EBCMVB_JMLR}.

\begin{figure}[t]
\centering
\includegraphics[width=.9\linewidth]{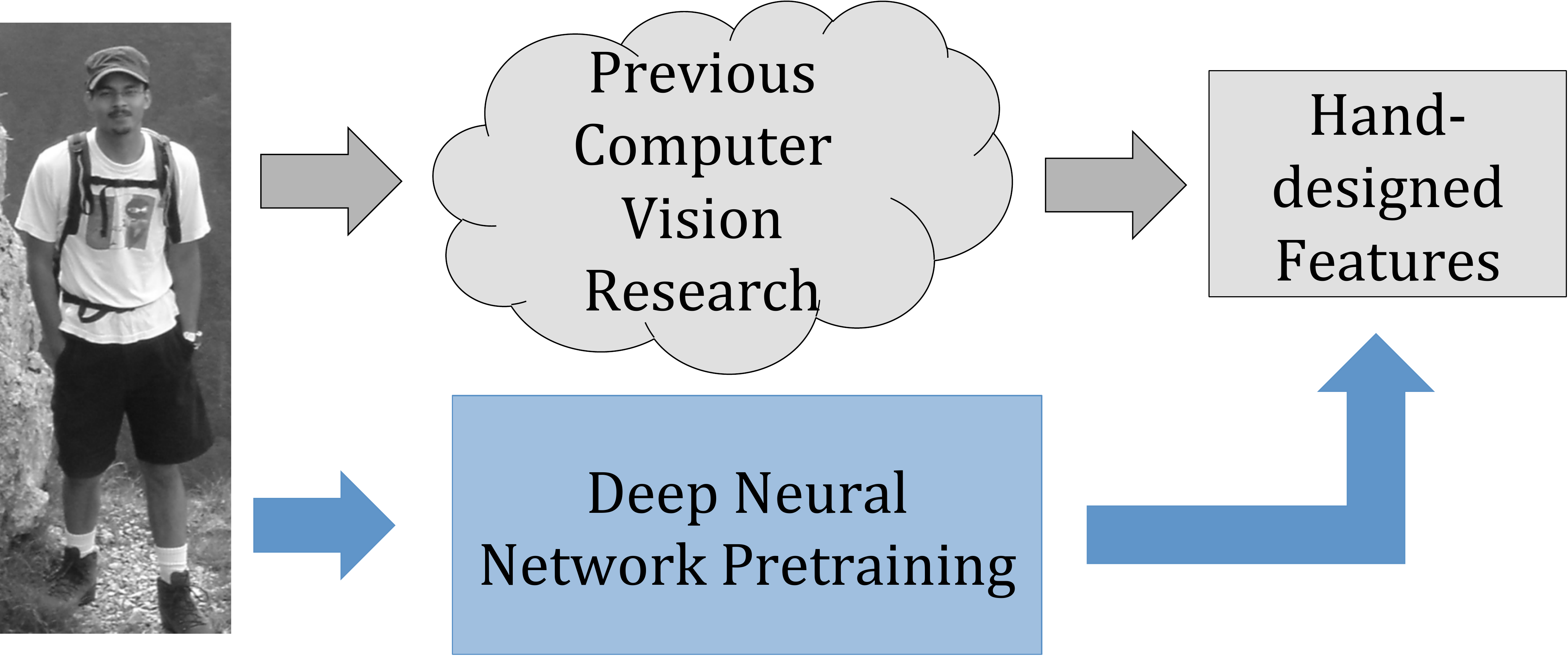}\\
\caption{{\bf Network pretraining via replicating hand-designed features:} We propose an unsupervised pretraining method based on hand-designed feature replication for deep learning. Through learning to replicate the features, the deep network utilizes the past computer vision research knowledge and learns a way to model the visual object structure. A subsequent supervised finetuning step further optimizes the deep network to achieve better recognition performance.}
\label{fig::intro}
\end{figure}

We propose to pretrain a deep network by solving a regression task that replicates the classic hand-designed feature extraction process (Fig.~\ref{fig::intro}). Classic hand-designed features such as the Histogram of Oriented Gradients (HOG)~\cite{Dalal_DT_CVPR05} and region covariance (COV)~\cite{Tuzel_TPM_ECCV06}, etc., are the results of years of research effort and human intelligence, and are established tools for modeling visual objects appearance. Our method is developed based on the insight that if the network can learn to replicate the hand-designed features, then it learns a way to capture object structure, similar to the hand-designed features. The following finetuning step can further boost the recognition performance by incorporating class specific information. Although using hand-designed features, our method is in line with the feature learning paradigm since it learns representations from data automatically. The hand-designed features are only used as ``guidance'' for initializing the deep network. Furthermore, our proposed pretraining method is unsupervised since the hand-designed features are extracted and used for pretraining without label information.

For certain visual object recognition tasks, the discriminative information is quite sparse as compared to the available information contained in the image. For example, shape is an important cue but color is not for pedestrian detection. The reconstruction-based pretraining methods have the tendency of favoring encoding information required for reconstructing the image in the network, which dilutes the discriminative information critical for the task in the network. When such discriminative information is present in the human-designed feature, our method encourages the network to encode this discriminative information primarily.

We verify the proposed pretraining method on the pedestrian detection task, for which many hand-designed features have been developed. In particular, the HOG and region covariance features have achieved remarkable performance and extensive use. We apply the proposed method to pretrain two convolutional neural networks (CNNs), where one replicates the HOG feature while the other replicates the region covariance feature. Our feature replication networks are then finetuned with class label information. We evaluate the performance on two public datasets. The experimental results show that the proposed method achieves better performance than the baseline methods.

\subsection{Contributions}

The contributions of the paper are listed below:
\begin{itemize}
\item We propose an unsupervised pretraining method based on replicating the process of hand-designed feature extraction for training a neural network.
\item We apply the proposed method to pretrain two neural networks, one replicating the process of HOG feature extraction and the other replicating the process of region covariance feature extraction, for the pedestrian detection task. Experimental results show that our pretraining method achieves favorable performance compared to several autoencoder-based pretraining methods. The resulting network achieves near state-of-the-art performance on pedestrian detection problem.
\end{itemize}

\section{Related Work}

For a long time, deep neural networks were believed to be too hard to train due to the tendency of gradient-based backpropagation methods to get stuck in local minima or flat regions, starting from random initializations. First effective strategies for unsupervised pretraining of deep networks were based on greedy layer-wise optimization such as Restricted Boltzmann Machines (RBMs) \cite{hinton2006fast}, and Stacked Autoencoders \cite{Ranzato_RBL_NIPS,Bengio_BLPL_NIPS}. RBMs operate based on an energy-minimization criterion over the training set. The autoencoder tries to reconstruct the output using the activations of the hidden layer, often with the aim of learning a compressed representation of the input where the number of hidden units is less than the number of inputs and layer sizes are decreasing. Autoencoders have also been shown to lead to networks with good generalization performance when the layer sizes are non-decreasing \cite{Bengio_BLPL_NIPS}.

Unsupervised methods have also been developed for pretraining CNNs, involving the use of convolutional RBMs \cite{Lee_LGRN_ICML}, sparse-coding autoencoders \cite{Ranzato_RPCL_NIPS}, and sparse-coding based non-linear transforms \cite{Kavukcuoglu_KSBGML_NIPS,Kavukcuoglu_KRFL_CVPR,Sermanet_SKC_CVPR13}.

We verify our proposed pretraining method on the pedestrian detection task. Here, we briefly review popular techniques in this area. Hand-designed features have been extensively researched and used for pedestrian detection. Gradient-based features such as Haar-like features \cite{Viola_VJS_CVPR}, SIFT \cite{Vedaldi_VGVZ_ICCV}, and the very popular HOG \cite{Dalal_DT_CVPR05} have been successfully used for the task. Texture-based features such as the Local Binary Pattern (LBP) have been used along with HOG in \cite{Wang_WXY_ICCV}. More recently, Doll\'{a}r \emph{et al.} \cite{Dollar_DTPB_BMVC09} proposed Integral Channel Features based on gradient histograms, gradients, and the LUV color space. The current state-of-the-art \emph{Spatial Pooling} \cite{PSH_ECCV14} uses covariance features and LBP along with spatial pooling. Some other examples of hand-designed features for pedestrian detection include~\cite{Benenson_BMTG_CVPR13,Lim_LZD_CVPR13,Mathias_MBTG_ICCV13}.

There were several deep learning-based methods for pedestrian detection. Sermanet \emph{et al.} \cite{Sermanet_SKC_CVPR13} use a sparse-coding based method \cite{Kavukcuoglu_KSBGML_NIPS} to learn a dictionary of filters in an unsupervised manner and \cite{Ouyang_WW_ICCV13} learns a deep-network with added deformation and occlusion handling. The success of human-designed features over deep learning methods for pedestrian detection further supports our belief that past human experience in the form of hand-designed features can guide learning deep networks.

%The paper is organized as follows: In Section \ref{sec::pretraining}, we present the proposed pretraining method. We illustrate its application to the pedestrian detection task in Section \ref{sec::pedestrian}. Experimental results are shown in Section \ref{sec::expr} and Section \ref{sec::conc} concludes the paper.

\section{Pretraining via Feature Replication}\label{sec::pretraining}

\begin{figure*}[t]
\centering
\includegraphics[width=.85\linewidth]{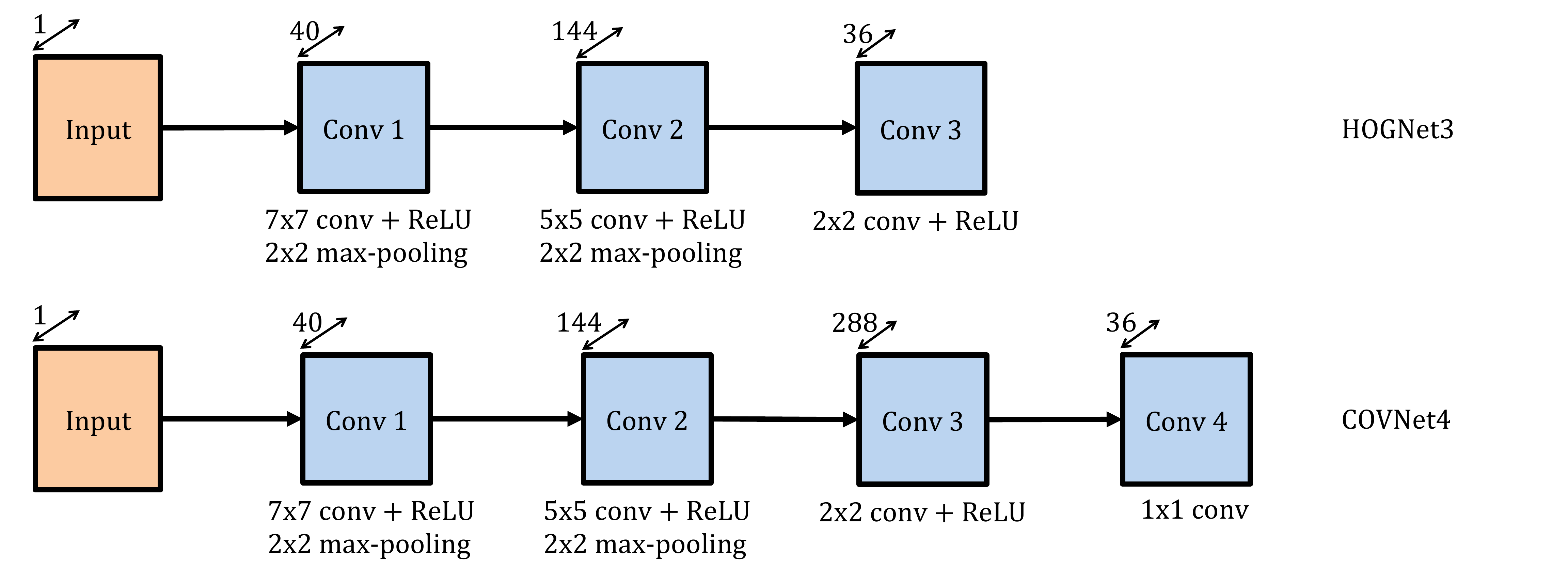}\\
\caption{{\bf Experimental network structures:} The HOGNet3 \& COVNet4 networks are pretrained through Euclidean loss minimization between the network output and the HOG \& COV features of the input image, respectively.}
\label{fig::REG_NETWORK}
\end{figure*}

Our pretraining method is based on the replication of hand-designed features. While various hand-designed features can be used for the task, in this paper, we consider two examples: the HOG feature and the region covariance feature. For pretraining by replicating the HOG feature, we use a three-layer convolutional neural network, and for the region covariance feature, we use a four layer convolutional neural network. Note that different network architectures can be used to replicate these features. The only requirements are that the number of output units should match the feature dimension, and the range of the output units should cover the range of the features (non-negative or real values).

\subsection{HOG Feature Replication}
 
We use a three-layer CNN shown in Fig.~\ref{fig::REG_NETWORK} in our experiments and pretrain it using the HOG feature \cite{Dalal_DT_CVPR05}. We refer to this network as \textrm{HOGNet3}. HOG feature computation utilizes a hierarchy of primitives of different levels of granularity. From high to low, they are image, block, cell, and pixel. An image contains a set of overlapping blocks, each block contains several cells, and each cell contains several pixels. The feature computation consists of three steps: 1) pixel-wise gradient computation, 2) cell-wise oriented gradient histogram computation, and 3) block-wise histogram normalization and concatenation. Specifically, the image is first convolved with derivative masks in the $x$ and $y$ directions for obtaining pixel-wise gradient magnitude and orientation. This is followed by quantizing gradient orientation to a set of bins, evenly spaced over 0 to 180 degrees. Pixels in a cell then distribute their gradient magnitudes to the orientation bins based on the gradient orientation. This constructs a histogram of oriented gradients for each cell. Histograms of the cells are individually normalized by the L2-norm of element-wise sum of cell histograms in the block. The normalized cell histograms are concatenated in a vector, which is the HOG feature for the block. The HOG features for the overlapping blocks in the image are further concatenated in a vector, which is the HOG feature for the image.

The power of the HOG feature is many-fold. The use of gradient orientation captures structural information of the object. The histogram binning avoids modeling unreliable fine details. The histogram normalization improves its invariance to illumination change. The HOG feature is generic in the sense that it does not contain class-specific information although better recognition performance can be achieved by using class-specific parameterization \cite{Felzenszwalb_FGMR_TPAMI10}. The orientation binning and histogram normalization are the two major sources that contribute to the HOG features' non-linear dependency on the input image.

We extract HOG features from grayscale image patches using the following parameters: image patch size of $16\times16$, $9$ orientation bins, cell-size of $8\times8$, and a block-size of $16\times16$. An image patch is mapped to a 36-dimensional HOG vector. The numbers of filters in the CNN layers are 40, 144, and 36 with sizes $7\times7$, $5\times5$, and $2\times2$, respectively. The number of filters in the last layer is 36 for matching the dimension of HOG features. In \textrm{HOGNet3}, each of the convolutional filter banks is followed by a bank of ReLUs for increasing the network's non-linearity, while keeping the optimization tractable. We apply non-overlapping max-pooling after the ReLU banks of the first and second CNN layers, with a pooling kernel size of $2\times2$. Due to the ReLU, outputs of \textrm{HOGNet3} are non-negative, the same as the HOG feature.

Our \textrm{HOGNet3} is trained via solving a multi-dimensional regression problem. The image patches are sampled from natural images and Euclidean loss is used as the regression objective function. We use back propagation and stochastic gradient descent with momentum to update the network parameters for minimizing the Euclidean loss.

\subsection{Region Covariance Feature Replication}

In the region covariance feature representation \cite{Tuzel_TPM_TPAMI08}, an image patch is represented by the covariance matrix of some pixel-wise features of the patch. While various pixel features can be used, we follow the original work \cite{Tuzel_TPM_TPAMI08} and use an eight-dimensional feature vector for each pixel. The eight dimensions are
\begin{equation}
 \mathbf{q} = \big{[}x,\medspace y,\medspace |I_x|,\medspace |I_y|,\medspace \sqrt{I_x^2 + I_y^2},\medspace |I_{xx}|,\medspace |I_{yy}|,\medspace \tan^{-1}\frac{I_x}{I_y}  \big{]}^T
 \nonumber%\label{eqn::covariance_feature}
\end{equation}
where $x$ and $y$ are the pixel coordinates, $I_x$ and $I_y$ are the first derivatives in $x$ and $y$ directions, and $I_{xx}$ and $I_{yy}$ are the second derivatives in $x$ and $y$ directions. Let $C$ denote the covariance matrix of $\mathbf{q}$ extracted from the pixels in an image region. Each of the elements in $C$ encodes the correlation between two dimensions of $\mathbf{q}$.

The covariance matrix $C$ has interesting properties that are useful for modeling the visual object appearance in an image region. Firstly, it is invariant to global illumination shift. Increasing or decreasing the image intensity value by a constant does not affect $C$. One difficulty in working with covariance features is that the space of positive definite (covariance) matrices is not a Euclidean space, but its structure can be better analyzed using elements of Riemannian geometry. To this end, we use the log-Euclidean geometry proposed in~\cite{arsigny2007geometric} and embed the manifold to a Euclidean space (tangent space) via the log transform: $\Sigma = U \log (S) U^T$ where $C=USU^T$ is the singular value decomposition of $C$. Note that the size of $\Sigma$ is the same as $C$ and is $8\times8$. Since $\Sigma$ is symmetric, we concatenate the upper triangular part of $\Sigma$ to form a compact vector representation for the region covariance feature, which is 36-dimensional. The region covariance feature can be extracted from an image region of any size. In this paper, we use a region size of $16\times16$.

We use a four-layer CNN (shown in Fig.~\ref{fig::REG_NETWORK}), referred to as \textrm{COVNet4}, and pretrain it using the region covariance feature. The first two layers of the \textrm{COVNet4} are the same as the \textrm{HOGNet3}. They are 40 and 144 convolutional filter banks of size $7\times7$ and $5\times5$ resp., followed by ReLUs and $2\times2$ non-overlapping max-pooling. The third layer consists of 288 filters of size $2\times2$, while the fourth layer has 36 filters of size $1\times1$. We insert a bank of ReLUs after the third convolutional filter bank but not after the fourth one. This is because the fourth layer is the output layer for replicating the region covariance feature, which is a real number vector. Similar to the \textrm{HOGNet3}, the \textrm{COVNet4} is trained via solving a multi-dimensional regression problem by minimizing the Euclidean loss.

\section{Pedestrian Detection}\label{sec::pedestrian}

We verify the proposed pretraining method on the pedestrian detection task. We first pretrain the \textrm{HOGNet3} and \textrm{COVNet4} to replicate the HOG and region covariance features respectively. They are then finetuned using labeled data for pedestrian detection.

We use the scanning window approach proposed in \cite{Viola_VJS_CVPR}, where a window is defined as a rectangular image region. For detecting and localizing pedestrians in an image, a classification model based on the window size is repeatedly applied to different image locations. It declares a detection at an image location if the classifier ouputs a score that is greater than a preset threshold. The image is scaled to form an image pyramid for detecting pedestrians of different sizes in the original image. The window size for our pedestrian hypothesis is set to $64\times128$, which is also the choice reported in the original hand-designed feature works \cite{Dalal_DT_CVPR05,Tuzel_TPM_TPAMI08}. For this window size, both \textrm{HOGNet3} and \textrm{COVNet4} produce an output feature map of size $13\times29\times36$ when the stride of the convolution filters is set to $1$.

We use the training images in the INRIA dataset \cite{Dalal_DT_CVPR05} for training the deep networks. The training set contains 614 images with humans and 1218 images without humans. In the original INRIA dataset, only some of the human locations are provided. We follow the approach described in \cite{Sermanet_SKC_CVPR13} and label all the human locations in the training set. This enables us to utilize the images with humans for hard-data mining, which is important as many false positive instances occur from parts of human bodies. We convert all the images to grayscale since both of the hand-designed features considered in the paper are designed for grayscale images. We thus only use the grayscale images for learning the deep networks.

We randomly sample a set of $16\times16$ image patches from the training images for pretraining. We extract the HOG and region covariance features using their respective feature extractors and train the \textrm{HOGNet3} and \textrm{COVNet4} to learn the relationships between the input image patch and the extracted features. The number of image patches used for pretraining is about 15 million. 

After pretraining, we finetune the networks using labeled data. We extract the pedestrian windows that are not heavily occluded from the training set for creating the positive sample set. In order to increase the variation, we perform data augmentation. We scale up and down the pedestrian windows by a factor of 1.05 and 0.95, respectively. We also randomly shift the pedestrian windows in the $x$ and $y$ directions by $2\%$ of their height. The pedestrian windows obtained from the training images are of different sizes. We scale the height to 128 pixels and use the method suggested in \cite{Dollar_DWSP_TPAMI08} to normalize the windows. The normalized windows are then scaled to a fixed size of $64\times128$. These constitute a set of positive samples which are around $90K$ in number. Also, we sample a set of windows from the training images that do not contain humans to construct the negative sample set. The number of samples of the negative set is around $50K$. These windows are resized to $64\times128$ to match the positive windows. We add a softmax output layer on top of the \textrm{HOGNet3} and \textrm{COVNet4} respectively and train the networks by minimizing the softmax loss. Below, we discuss some commonly used techniques for optimizing performance in object detection, which we integrate into our system.

Hard-data mining is a practical technique that is known to boost detection performance of the classifiers. After finetuning, we apply the sliding window technique and use the learned networks to detect pedestrians in all the images in the INRIA training set. We store the scores of the false positive windows and add the top scoring $20K$ false positive windows to the negative training set. We then finetune \textrm{HOGNet3} and \textrm{COVNet4}, where the network parameters are initialized using the parameters learned from the previous finetuning step. This completes a run of hard-data mining. We repeat the hard-data mining procedure for several runs until the number of false positive samples falls below $2K$. Overall, we perform 8 hard-data mining runs for the \textrm{HOGNet3} and 6 runs for \textrm{COVNet4}.

After hard-data mining, we remove the softmax output layer from the \textrm{HOGNet3} and \textrm{COVNet4} and use the networks as a feature extractor. The extracted features are then fed to a linear SVM for learning the maximum margin classifier. This approach has been adopted in several previous works including \cite{Girshick_GDDM_CVPR14,Sermanet_SKC_CVPR13}. Instead of using the L1-Hinge Loss SVM, we use the L2-Hinge Loss SVM as the quadratic cost allows for faster optimization. To train the SVM, we use \emph{LIBLINEAR}, a publicly available large-scale linear SVM implementation ~\cite{Fan_FCHWL_JMLR08}. We use the training data obtained from the last hard-data mining iteration for training the SVM, where the negative data is weighted 20 times more than the positive data and the L2-Hinge Loss is weighted by 1.

The sliding window detection approach often declares several detections around a pedestrian hypothesis. In order to combine the multiple detections, we apply non-maximum suppression. We use the pairwise window matching method for non-maximum suppression. Let $W_1$ and $W_2$ be two window hypotheses and assume that the detection score of $W_1$ is larger than that of $W_2$. We suppress $W_2$ if it has a large overlap with $W_1$. Two popular measures are available for measuring the overlap. One is based on the intersection over union measure (IoU), which is the ratio of the size of the intersection of $W_1$ and $W_2$ divided by the size of their union. The other measure is based on the ratio of the size of the intersection of $W_1$ and $W_2$ over the size of $W_2$, called Io2. We found that applying the two measures in sequence renders slightly better performance than applying either of them alone for our detector networks. The threshold we use for IoU is 0.4, while the threshold for Io2 is 0.6.

In order to reduce localization errors, we train a linear regression model to improve the bounding boxes of the pedestrian detections obtained from the SVM classifier. Following the methodology of~\cite{Girshick_GDDM_CVPR14}, we use the features outputted from the last layer of the network to learn four functions that map the $x$, $y$ locations, height, and width of the proposed detections ($P^i$) to their corresponding ground truth locations ($G^i$) using pairs ($P^i, G^i$). The ground truth pairs are collected from applying the SVM classifier to scan through the pedestrian hypothesis in the training set. During test time, we apply the learned functions on the proposed detections ($P^j$) to map them to a predicted ground-truth box ($\hat{G}^j$). The training data ($P^i, G^i$) is selected such that $P^i$ has an SVM score of at least $0.5$ and intersects $G^i$ with an IoU measure of at least $0.6$.

\section{Experiments}\label{sec::expr}

We first analyze the proposed pretraining method for replicating two hand-designed features. Next, we finetune the networks for the pedestrian detection task and compared them with the existing algorithms. All the experiments were conducted using a PC with an Intel i7 multi-core processor and a Nvidia K40 Tesla GPGPU card.

\subsection{Experimental Results for Pretraining}

We used the publicly available convolutional neural network library \emph{Caffe} \cite{jia2014caffe} for training the deep networks. It utilizes the GPU architecture to parallelize the feedforward and gradient computations, allowing efficient training of the network. The mini-batch size used for stochastic gradient descent was $10.5K$, with a base learning rate of 0.001 and momentum weight of 0.9. As the magnitude of the region covariance feature was larger than the HOG feature, we used a smaller base learning rate of 0.00001. We used the inverse law for adjusting the learning rate for pretraining. The networks were regularized by the L2 loss on the network parameters with a weight decay of 0.0005. As described in the section~\ref{sec::pedestrian}, the training data was obtained from the INRIA training set.

In order to evaluate the pretraining performance, we also sampled a set of image patches from the INRIA test dataset. Similar to the INRIA training dataset, the INRIA test dataset consists of two sets of images: one set with humans and the other without humans. The number of images in the two sets are 287 and 453, respectively. We converted all the images in the dataset to grayscale and randomly sampled 10 million $16\times16$ image patches from both the sets and used the HOG and region covariance feature extractors to construct the test set.

We pretrained the \textrm{HOGNet3} and \textrm{COVNet4} by minimizing the Euclidean loss between the network outputs and the extracted hand-designed features. We trained the \textrm{HOGNet3} for 100 epochs but trained the \textrm{COVNet4} for 40 epochs since the training loss decreased very slowly after 30 epochs for \textrm{COVNet4}. After each epoch, we stored the parameters of the networks and evaluated their performance on the test set.

\begin{figure}[t]
\centering
\includegraphics[width=.99\linewidth]{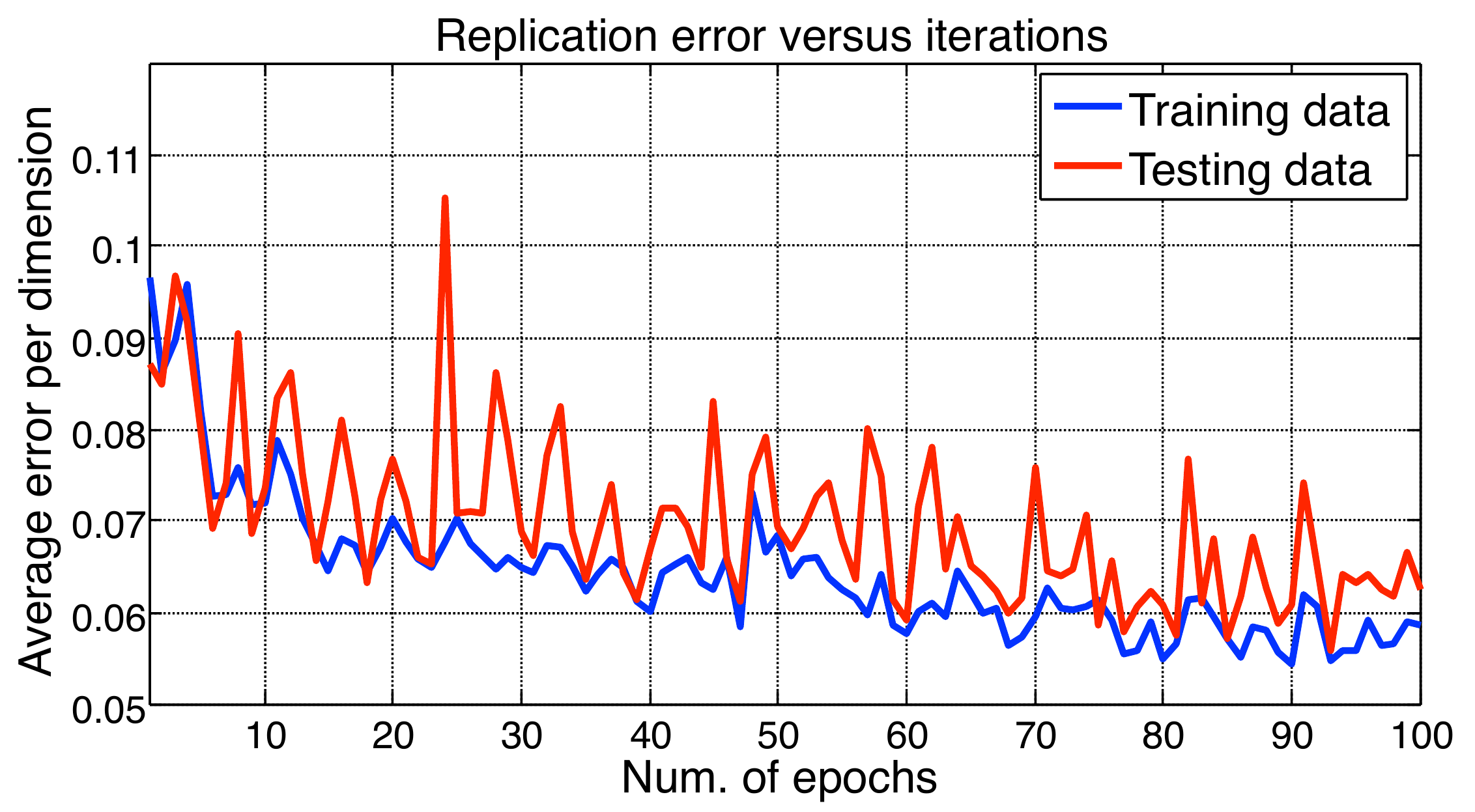}
\caption{{\bf Replication of the HOG features:} The figure plots the training and test error of the \textrm{HOGNet3}.}
\label{fig::HOG_PRETRAINING}
\end{figure}

\begin{figure}[t]
\centering
\includegraphics[width=.99\linewidth]{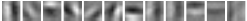}
\caption{{\bf Visualization of \textrm{HOGNet3}:} The figure shows some of the first layer convolutional filters in the \textrm{HOGNet3}. The filters resemble edge detectors of different orientations, capturing the behavior of the HOG feature extractor.}
\label{fig::HOG_Filters}
\end{figure}

In Fig.~\ref{fig::HOG_PRETRAINING}, we plot the pretraining performance of the \textrm{HOGNet3} where the $y$-axis is the average estimation error per feature dimension and the $x$-axis denotes the number of epochs. As shown in the graph, both of the training errors decreased gradually with some fluctuations. We did not observe significant overfitting. After 100 epochs, the average error was reduced to 0.065 on the test set, where the average magnitude of the HOG features was 0.14.% in the test set.

Fig.~\ref{fig::HOG_Filters} visualizes the filters learned in the first convolutional layer of the \textrm{HOGNet3}. We found that the filters resembled edge filters of different orientations. As the HOG feature is based on oriented gradients, these filters captured gradient magnitudes at different orientations from the input image and passed the results to the succeeding layers for mimicking the histogram operation. The remaining filters not shown either resembled edge filters of small gradients or some small magnitude random patterns. We did not observe filters resembling texture filters as observed in \cite{Sermanet_SKC_CVPR13}.

\begin{figure}[t]
\centering
\includegraphics[width=.99\linewidth]{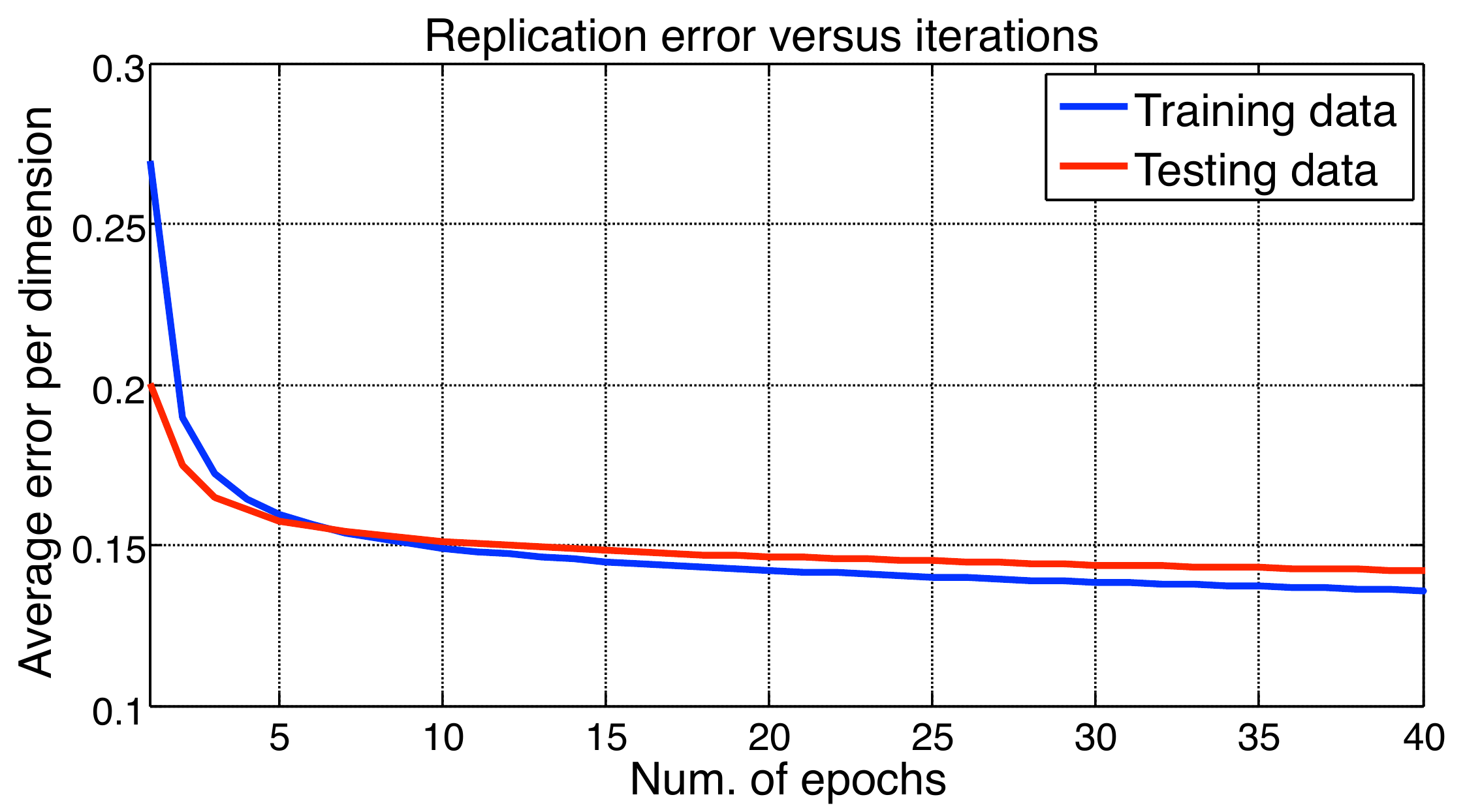}
\caption{{\bf Replication of the region covariance features:} The figure plots the training and test error of the \textrm{COVNet4}.}
\label{fig::COV_PRETRAINING}
\end{figure}

In Fig.~\ref{fig::COV_PRETRAINING}, we plot the pretraining performance of replicating the region covariance features using the \textrm{COVNet4}. Unlike the replication of the HOG features, we found that both of the training and test errors decreased smoothly, probably due to the smaller learning rate. After 40 epochs, the testing error per feature dimension was about 0.14, while the average magnitude of the region covariance feature was 0.26.

\begin{figure}[t]
\centering
\includegraphics[width=.99\linewidth]{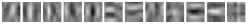}
\caption{{\bf Visualization of \textrm{COVNet4}:} The figure shows some of the first layer convolutional filters in the \textrm{COVNet4}. The filters resemble $1^{st}$ and $2^{nd}$ order image derivative operators, capturing the behavior of the region covariance feature extractor.}
\label{fig::COV_Filters}
\end{figure}

Fig.~\ref{fig::COV_Filters} displays the first layer convolutional filters learned by replicating the region covariance features. Unlike those in \textrm{HOGNet3}, the filters in \textrm{COVNet4} were more similar to the first-order and second-order gradient operators in the $x$ and $y$ directions, which are used to compute the region covariance features.% in Equation~(\ref{eqn::covariance_feature}). Similar to the HOG case, the remaining filters were not texture filters but random patterns of small magnitude.

\subsection{Experimental Results for Pedestrian detection}

We finetuned the \textrm{HOGNet3} and \textrm{COVNet4} using the labeled data in the INRIA training set for pedestrian detection as described in Section~\ref{sec::pedestrian}. As the pedestrian window size was several times larger than the image patch used in the pretraining, we decreased the the mini-batch size to 100 in order to fit the data into the memory. The base learning rate was set to 0.01 for both networks, while the other parameters were kept the same as those used in the pretraining step. We initialized the softmax-appended \textrm{HOGNet3} and \textrm{COVNet4} by using the network parameters learned in the respective pretraining steps and finetuned them for 10 epochs. A dropout layer \cite{srivastava2014dropout} was added to the last feature layer for better generalization performance. After finetuning, we applied the sliding window technique for hard-data mining using the INRIA training set as described in Section~\ref{sec::pedestrian}.

For facilitating the comparison, we also applied the autoencoder technique to pretrain a deep network using a similar architecture. We used an autoencoder which tries to reconstruct the half-sized input through a fully-connected layer after three convolutional layers of dimensions same as the \textrm{HOGNet3} network. We trained the network to reconstruct the input in the end-to-end fashion instead of using the greedy layerwise pretraining. This network is referred to as \textrm{AutoNet3}. Further, we also finetuned a network with the same architecture as \textrm{HOGNet3} but initialized with random weights to show the usefulness of our pretraining method.

\begin{figure}[t]
\centering
\includegraphics[width=.99\linewidth]{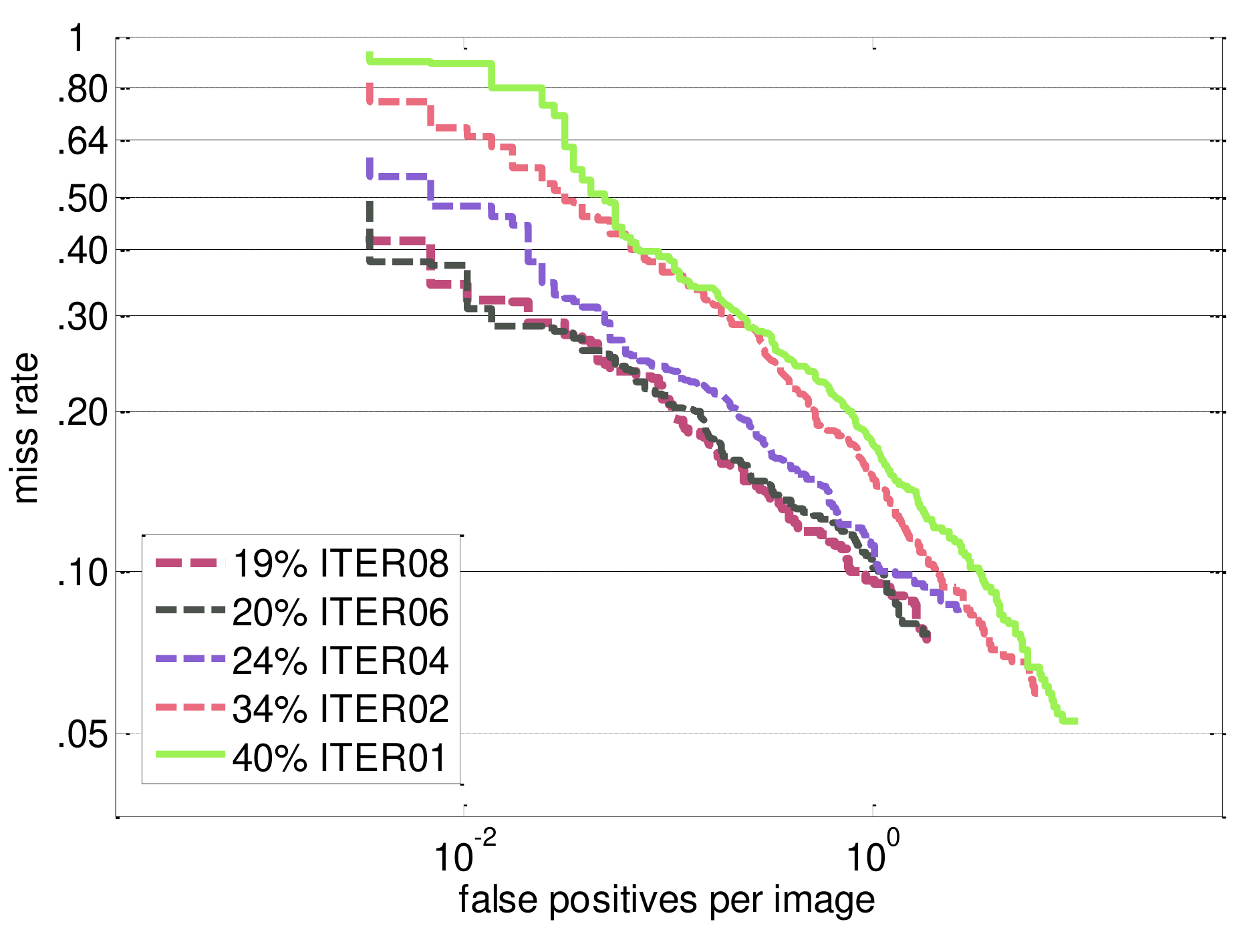}\\
\caption{{\bf Effect of hard-data mining:} The figure illustrates the detection performance improvement on the INRIA test dataset, from the hard-data mining step performed during finetuning the \textrm{HOGNet3}.}
\label{fig::hard_data}
\end{figure}
\begin{figure}[t]
\centering
\includegraphics[width=.99\linewidth]{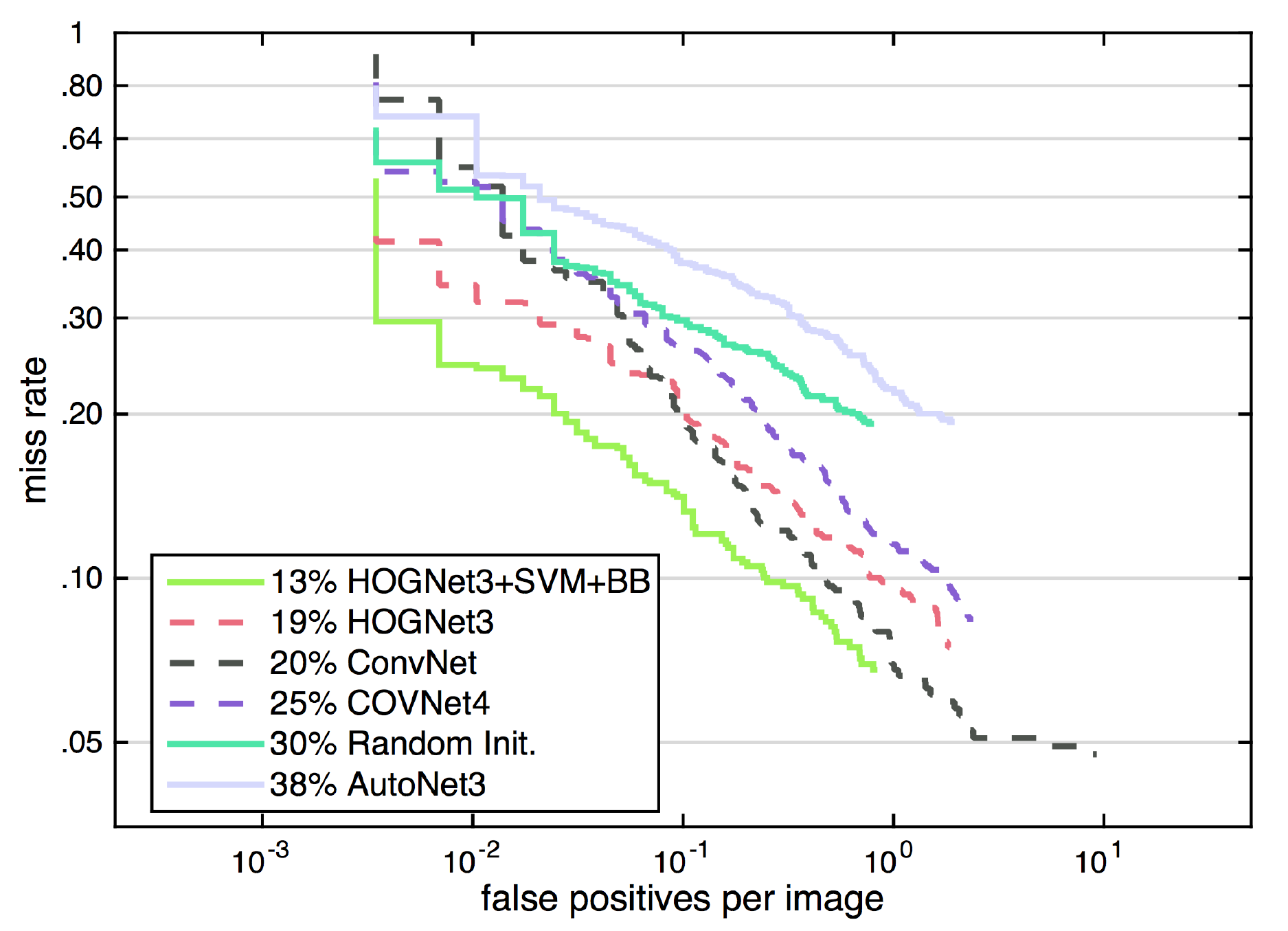}\\
\caption{{\bf Pretraining method comparison:} The figure compares the performance of the networks trained with different pretraining methods on the INRIA test dataset. Our \textrm{HOGNet3} achieves the best performance, outperforming the \textrm{ConvNet} \cite{Sermanet_SKC_CVPR13}, which is trained based on a sparse-coding based pretraining method~\cite{Kavukcuoglu_KSBGML_NIPS}.}
\label{fig::hogcov_comp}
\end{figure}

We evaluated the performance of the trained networks for pedestrian detection using the INRIA test dataset and Daimler test dataset. The INRIA test dataset, composed of personal photographs, contains 287 color images with humans, which we converted to grayscale. The Daimler test dataset was captured using a camera mounted on a vehicle driving around in a city and the captured images are in grayscale. The Daimler dataset, which contains 21787 images is several magnitudes larger than the INRIA dataset. To detect pedestrians at different scales, we searched through three octaves with a base scale of 1.07. Although the Daimler dataset also includes a training set, we did not retrain our networks using the set. Our networks were trained using the INRIA training set. %There are several other datasets proposed for pedestrian detection as summarized in~\cite{Dollar_DWSP_TPAMI08}, all of which are taken from street views in the color image format, except for the INRIA dataset and Daimler dataset. Our network does not utilize color information, which is a valuable cue for pedestrian detection since the colors of buildings, trees, and traffic signs are usually very different to a pedestrian's appearance.

The Caltech toolbox described in~\cite{Dollar_DWSP_TPAMI08} was used for performance evaluation. The detection performance of each algorithm was visualized using a Receiver Operating Characteristic (ROC) curve where the $y$ axis is the miss rate and the $x$ axis is the false positive per image (FPPI). The curve is plotted in the log scale. It is difficult to compare two algorithms based on the ROC curves as the toolbox uses the average miss rate for summarizing the ROC curve. Specifically, the average of the miss rates for the 9 points evenly spread between 0.01 to 1 FPPI in the log scale is used to summarize the performance of an algorithm. The miss rate from the region where FPPI is less than 0.01 is not used in the comparison.

In Fig.~\ref{fig::hard_data}, we show the performance improvement resulting from the hard-data mining step during finetuning of the \textrm{HOGNet3}. Similar behavior was displayed by the \textrm{COVNet4}. While more iterations helped, the improvements also diminished with iterations.

We compare the performance of the networks trained using different pretraining techniques in Fig.~\ref{fig::hogcov_comp}. It includes the \textrm{HOGNet3}, \textrm{COVNet4}, \textrm{AutoNet3}, \textrm{Random Init.}, and \textrm{ConvNet} \cite{Sermanet_SKC_CVPR13}. The \textrm{ConvNet} had a very different structure, using YUV images as input, and absolute
value rectification and contrast normalization for nonlinearity. It used a sparse-coding based pretraining method. %It also randomly selected $20\%$ of the neuron connections for breaking the symmetry and used an SVM for the output layer, along with a sparse-coding based pretraining.

From Fig.~\ref{fig::hogcov_comp}, we observe that the \textrm{HOGNet3} obtained an average miss rate of $19\%$, outperforming the $20\%$ of the the \textrm{ConvNet} \cite{Sermanet_SKC_CVPR13}, the $25\%$ of the the \textrm{COVNet4}, the $38\%$ of the \textrm{AutoNet3} and the $30\%$ of the randomly initialized network. When utilizing the SVM classifier and the bounding box prediction technique, our performance was further improved to $13\%$. The \textrm{AutoNet3} did not perform well since it tried to encode the information required for reconstruction, which dilutes the critical information for pedestrian detection. The improvement over random initialization of the network clearly shows the usefulness of our pretraining scheme. In the remaining part of the section, we only report the performance of the \textrm{HOGNet3} when used with SVM and the bounding box regression technique.

\begin{figure}[t]
\centering
\includegraphics[width=.99\linewidth]{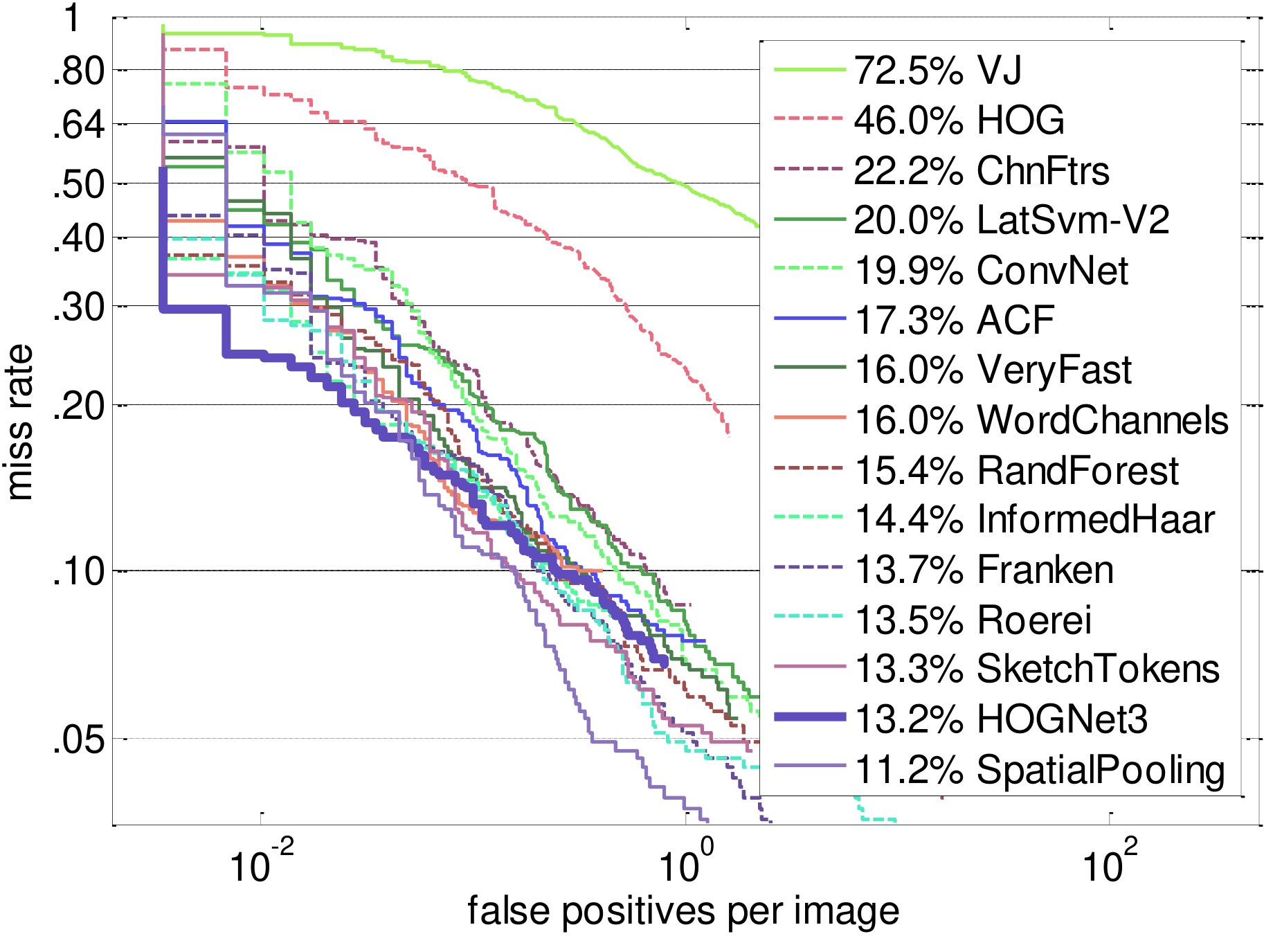}\\
\caption{{\bf Results on the INRIA test dataset:} }
\label{fig::inria_perm}
\end{figure}
\begin{figure}[t]
\centering
\includegraphics[width=.99\linewidth]{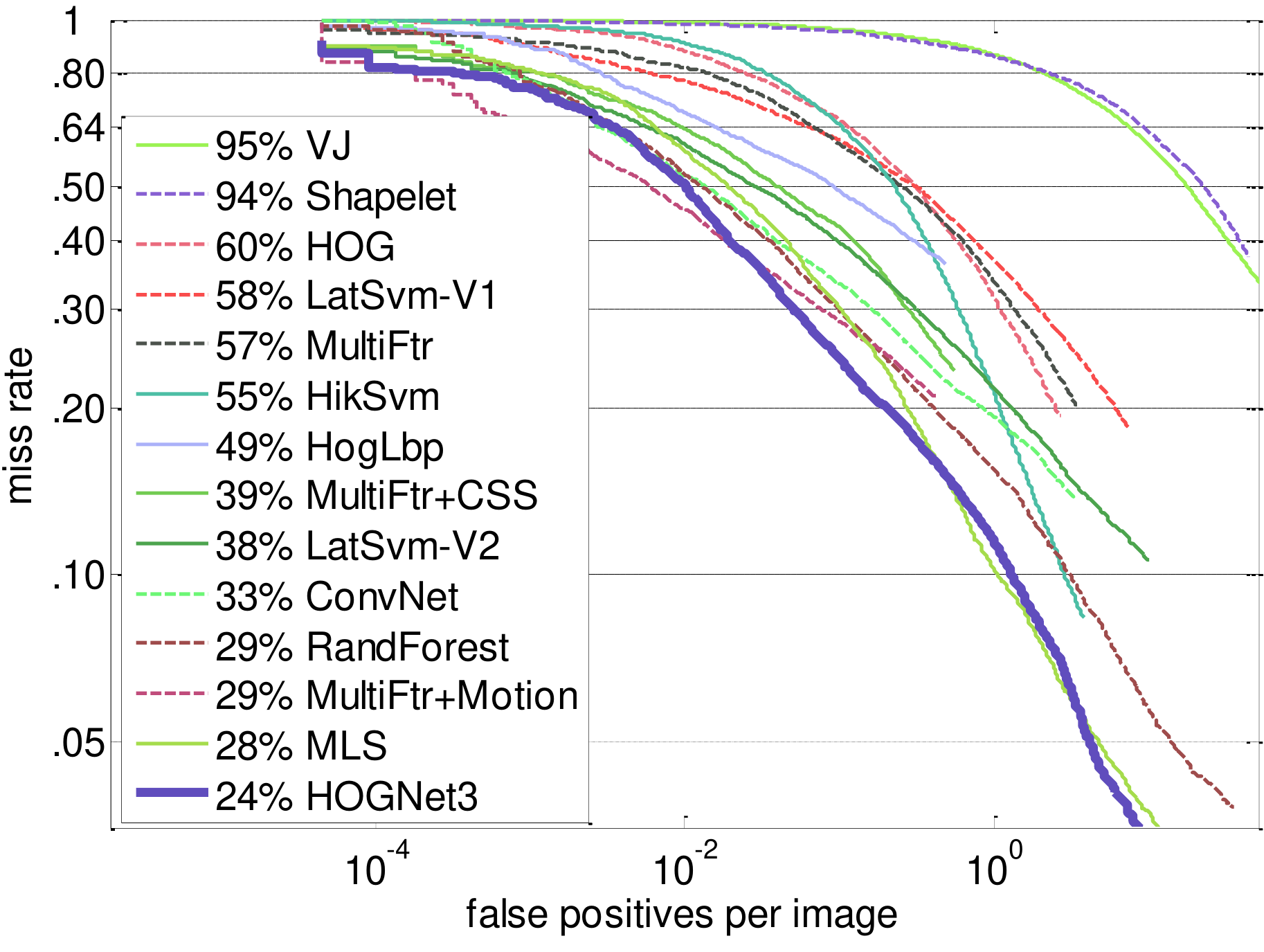}\\
\caption{{\bf Results on the Daimler test dataset:} }
\label{fig::daimler_perm}
\end{figure}
\begin{figure*}[t!]
\centering
\includegraphics[width=.99\linewidth]{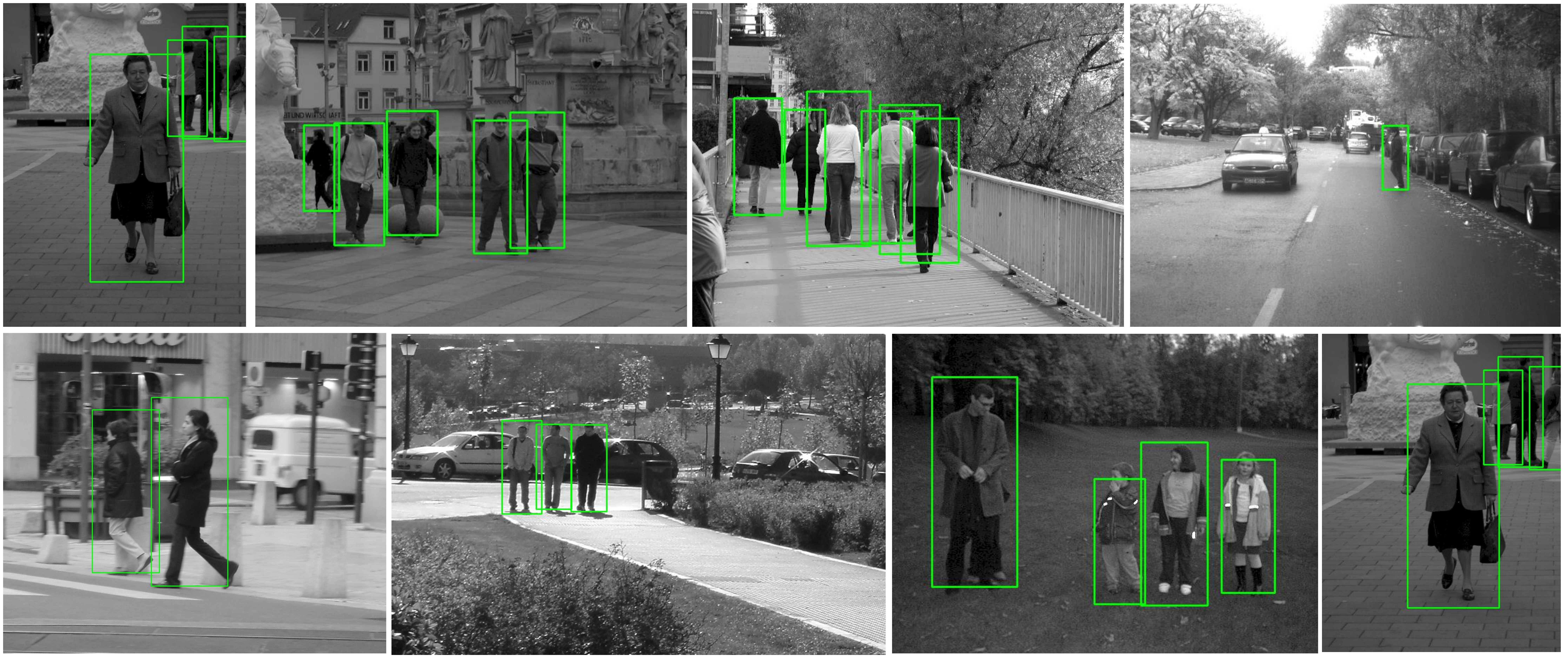}\\
\caption{Example pedestrian detection outputs of the \textrm{HOGNet3} on the INRIA and Daimler test datasets.}
\label{fig::pedestrians}
\end{figure*}

%had a very different structure, using YUV images as input, and absolute value rectification and contrast normalization for nonlinearity. It also randomly selected $20\%$ of the neuron connections for breaking the symmetry and used an SVM for the output layer, along with a sparse-coding based pretraining.

Fig.~\ref{fig::inria_perm} compares the \textrm{HOGNet3} with the state-of-the-art pedestrian detection methods on the INRIA test set. The performance data of the other algorithms were obtained from the Caltech pedestrian detection website. The \textrm{HOGNet3} obtained an average miss rate of $13.2\%$, which was second only to the \emph{Spatial Pooling} algorithm \cite{PSH_ECCV14}. It significantly outperformed the baseline methods including the HOG detector ($46.0\%$) \cite{Dalal_DT_CVPR05} and \textrm{ConvNet} ($19.9\%$) \cite{Sermanet_SKC_CVPR13}.

%The Caltech toolbox only measured the detection performance between 0.01 to 1.0 FPPI. It excludes the low FPPI region which is of great interest for autonomous driving. When extending the FPPI range included for computing the average miss rate to the leftmost point of the curves, we found that {\bf the \textrm{HOGNet3} achieved the best average miss rate of 16.2\%}, which was better than the average miss rate of $16.9\%$ obtained by \cite{PSH_ECCV14}.

Fig.~\ref{fig::daimler_perm} compares the performance of the \textrm{HOGNet3} with the state-of-the-art pedestrian detection methods on the Daimler test set. The \textrm{HOGNet3} performed better than the \textrm{ConvNet} \cite{Sermanet_SKC_CVPR13} to obtain the first place. In Fig.~\ref{fig::pedestrians}, we visualize some of the pedestrian detection outputs on the INRIA and Daimler datasets.

\section{Conclusion}\label{sec::conc}

Through the example of pedestrian detection, we have shown that hand-designed features can be used for pretraining deep neural networks in a simple but effective way. Our method was based on the insight that discriminative information encoded in the hand-designed feature can be transferred to the network via pretraining. It can be later integrated with class specific information in the finetuning stage to boost the performance. Our method is useful for the recognition task where only a small amount of training data are available but a good feature can be hand engineered.

{\small
\bibliographystyle{ieee}
\bibliography{replicate}

\begin{thebibliography}{10}\itemsep=-1pt

\bibitem{arsigny2007geometric}
V.~Arsigny, P.~Fillard, X.~Pennec, and N.~Ayache.
\newblock Geometric means in a novel vector space structure on symmetric
  positive-definite matrices.
\newblock {\em SIAM journal on matrix analysis and applications},
  29(1):328--347, 2007.

\bibitem{Benenson_BMTG_CVPR13}
R.~Benenson, M.~Mathias, T.~Tuytelaars, and L.~V. Gool.
\newblock Seeking the strongest rigid detector.
\newblock In {\em Conference on Computer Vision and Pattern Recognition}, 2013.

\bibitem{Bengio_BLPL_NIPS}
Y.~Bengio, P.~Lamblin, D.~Popovici, and H.~Larochelle.
\newblock Greedy layer-wise training of deep networks.
\newblock {\em Neural Information Processing Systems (NIPS)}, 2007.

\bibitem{Dalal_DT_CVPR05}
N.~Dalal and B.~Triggs.
\newblock Histograms of oriented gradients for human detection.
\newblock In {\em Conference on Computer Vision and Pattern Recognition}, 2005.

\bibitem{Dollar_DTPB_BMVC09}
P.~Dollar, Z.~Tu, P.~Perona, and S.~Belongie.
\newblock Integral channel features.
\newblock In {\em British Machine Vision Conference (BMVC)}, 2009.

\bibitem{Dollar_DWSP_TPAMI08}
P.~Dollar, C.~Wojek, B.~Schiele, and P.~Perona.
\newblock Pedestrian detection: an evaluation of the state of the art.
\newblock {\em IEEE Transaction on Pattern Analysis and Machine Intelligence},
  34(4):743--761, 2012.

\bibitem{Erhan_EBCMVB_JMLR}
D.~Erhan, Y.~Bengio, A.~Courville, P.-A. Manzagol, P.~Vincent, and S.~Bengio.
\newblock Why does unsupervised pre-training help deep learning?
\newblock {\em Journal on Machine Learning Research}, 11:625--660, 2010.

\bibitem{Fan_FCHWL_JMLR08}
R.-E. Fan, K.-W. Chang, C.-J. Hsieh, X.-R. Wang, and C.-J. Lin.
\newblock Liblinear: a library for large linear classification.
\newblock {\em IEEE Transaction on Pattern Analysis and Machine Intelligence},
  9:1871--1874, 2008.

\bibitem{Felzenszwalb_FGMR_TPAMI10}
P.~F. Felzenszwalb, R.~B. Girshick, D.~McAllester, and D.~Ramanan.
\newblock Object detection with discriminatively trained part based models.
\newblock {\em IEEE Transaction on Pattern Analysis and Machine Intelligence},
  32(9):1627--1645, 2010.

\bibitem{Girshick_GDDM_CVPR14}
R.~Girshick, J.~Donahue, T.~Darrell, and J.~Malik.
\newblock Rich feature hierarchies for accurate object detection and semantic
  segmentation.
\newblock In {\em Conference on Computer Vision and Pattern Recognition}, 2014.

\bibitem{hinton2002training}
G.~Hinton.
\newblock Training products of experts by minimizing contrastive divergence.
\newblock {\em Neural computation}, 14(8):1771--1800, 2002.

\bibitem{hinton2006fast}
G.~Hinton, S.~Osindero, and Y.-W. Teh.
\newblock A fast learning algorithm for deep belief nets.
\newblock {\em Neural computation}, 18(7):1527--1554, 2006.

\bibitem{jia2014caffe}
Y.~Jia, E.~Shelhamer, J.~Donahue, S.~Karayev, J.~Long, R.~Girshick,
  S.~Guadarrama, and T.~Darrell.
\newblock Caffe: Convolutional architecture for fast feature embedding.
\newblock {\em arXiv preprint arXiv:1408.5093}, 2014.

\bibitem{Kavukcuoglu_KRFL_CVPR}
K.~Kavukcuoglu, M.~Ranzato, R.~Fergus, and Y.~Le-Cun.
\newblock Learning invariant features through topographic filter maps.
\newblock In {\em Conference on Computer Vision and Pattern Recognition}, 2009.

\bibitem{Kavukcuoglu_KSBGML_NIPS}
K.~Kavukcuoglu, P.~Sermanet, Y.~lan Boureau, K.~Gregor, M.~Mathieu, and Y.~L.
  Cun.
\newblock Learning convolutional feature hierarchies for visual recognition.
\newblock In {\em Neural Information Processing Systems (NIPS)}. 2010.

\bibitem{Krizhevsky_KSH_NIPS12}
A.~Krizhevshky, I.~Sutskever, and G.~E. Hinton.
\newblock Imagenet classification with deep convolutional neural netowrks.
\newblock In {\em Neural Information Processing Systems (NIPS)}, 2012.

\bibitem{Lee_LGRN_ICML}
H.~Lee, R.~Grosse, R.~Ranganath, and A.~Y. Ng.
\newblock Convolutional deep belief networks for scalable unsupervised learning
  of hierarchical representations.
\newblock In {\em International Conference on Machine Learning}, 2009.

\bibitem{Lim_LZD_CVPR13}
J.~J. Lim, C.~L. Zitnick, and P.~Dollar.
\newblock Sketch tokens: a learned mid-level representation for contour and
  object detection.
\newblock In {\em Conference on Computer Vision and Pattern Recognition}, 2013.

\bibitem{Mathias_MBTG_ICCV13}
M.~Mathias, R.~Benenson, R.~Timofte, and L.~V. Gool.
\newblock Handling occlusions with franken-classifiers.
\newblock In {\em International Conference on Computer Vision}, 2013.

\bibitem{Ouyang_WW_ICCV13}
W.~Ouyang and X.~Wang.
\newblock Joint deep learning for pedestrian detection.
\newblock In {\em International Conference on Computer Vision}, 2013.

\bibitem{PSH_ECCV14}
S.~Paisitkriangkrai, C.~Shen, and A.~van~den Hengel.
\newblock Strengthening the effectiveness of pedestrian detection with
  spatially pooled features.
\newblock In {\em European Conference on Computer Vision}. 2014.

\bibitem{ranzato2006efficient}
M.~Ranzato, C.~Poultney, S.~Chopra, Y.~L. Cun, et~al.
\newblock Efficient learning of sparse representations with an energy-based
  model.
\newblock In {\em Neural Information Processing Systems (NIPS)}, pages
  1137--1144, 2006.

\bibitem{Ranzato_RBL_NIPS}
M.~A. Ranzato, Y.~lan Boureau, and Y.~L. Cun.
\newblock Sparse feature learning for deep belief networks.
\newblock In {\em Neural Information Processing Systems (NIPS)}. 2008.

\bibitem{Ranzato_RPCL_NIPS}
M.~A. Ranzato, C.~Poultney, S.~Chopra, and Y.~L. Cun.
\newblock Efficient learning of sparse representations with an energy-based
  model.
\newblock In {\em Neural Information Processing Systems (NIPS)}. 2007.

\bibitem{Sermanet_SKC_CVPR13}
P.~Sermanet, K.~Kavukcuoglu, and S.~Chintala.
\newblock Pedestrian detection with unsupervised multi-stage feature learning.
\newblock In {\em Conference on Computer Vision and Pattern Recognition}, 2013.

\bibitem{srivastava2014dropout}
N.~Srivastava, G.~Hinton, A.~Krizhevsky, I.~Sutskever, and R.~Salakhutdinov.
\newblock Dropout: A simple way to prevent neural networks from overfitting.
\newblock {\em Journal on Machine Learning Research}, 15(1):1929--1958, 2014.

\bibitem{Taigman_TYRW_CVPR14}
Y.~Taigman, M.~Yang, M.~A. Ranzato, and L.~Wolf.
\newblock Deepface: closing the gap to human-level performance in face
  verification.
\newblock In {\em Conference on Computer Vision and Pattern Recognition}, 2014.

\bibitem{Tuzel_TPM_ECCV06}
O.~Tuzel, F.~Porikli, and P.~Meer.
\newblock Region covariance: A fast descriptor for detection and
  classification.
\newblock In {\em European Conference on Computer Vision}. 2006.

\bibitem{Tuzel_TPM_TPAMI08}
O.~Tuzel, F.~Porikli, and P.~Meer.
\newblock Pedestrian detection via classification on riemannian manifolds.
\newblock {\em IEEE Transaction on Pattern Analysis and Machine Intelligence},
  2008.

\bibitem{Vedaldi_VGVZ_ICCV}
A.~Vedaldi, V.~Gulshan, M.~Varma, and A.~Zisserman.
\newblock Multiple kernels for object detection.
\newblock In {\em International Conference on Computer Vision}, 2009.

\bibitem{Viola_VJS_CVPR}
P.~Viola, M.~J. Jones, and D.~Snow.
\newblock Detecting pedestrians using patterns of motion and appearance.
\newblock In {\em International Conference on Computer Vision}, 2003.

\bibitem{Wang_WXY_ICCV}
X.~Wang, T.~Han, and S.~Yan.
\newblock An hog-lbp human detector with partial occlusion handling.
\newblock In {\em International Conference on Computer Vision}, 2009.

\end{thebibliography}
}

\end{document}